%% file: main.tex
\documentclass{article}

\usepackage{ijcai19}

\usepackage{times}
\usepackage{soul}
\usepackage{url}
\usepackage[hidelinks]{hyperref}
\usepackage[utf8]{inputenc}
\usepackage[small]{caption}
\usepackage{booktabs}
\usepackage{amssymb}
\usepackage{bm}
\usepackage{subfig}
\usepackage{amsmath}
\usepackage{mathabx}
\usepackage{graphicx}
\usepackage{xcolor}

\newcommand\intd{\mathop{}\mathrm{d}}
\DeclareMathOperator{\diag}{diag}
\DeclareMathOperator{\GammaText}{Gamma}
\DeclareMathOperator*{\argmax}{argmax}
\newcommand{\bomega}{\bm{\omega}}
\newcommand{\citet}[1]{\citeauthor{#1}~\shortcite{#1}}
\newcommand{\citep}{\cite}

\makeatletter
\newcommand\footnoteref[1]{\protected@xdef\@thefnmark{\ref{#1}}\@footnotemark}
\makeatother

\usepackage[nameinlink,capitalize]{cleveref}
\allowdisplaybreaks

\usepackage{xargs}

\graphicspath{{images/}{../images/}}

\title{Efficient Non-parametric Bayesian Hawkes Processes}

\author{
	Rui Zhang$^{1,2}$
	\and
	Christian Walder$^{1,2}$\and
	Marian-Andrei Rizoiu$^{3}$\And
	Lexing Xie$^{1,2}$
	\affiliations
	$^1$The Australian National University, Australia \\
	$^2$Data61 CSIRO, Australia \\
	$^3$University of Technology Sydney, Australia
	\emails
	$\left\{\text{firstname}\right\}$.$\left\{\text{lastname}\right\}$@[anu.edu.au$^1$, data61.csiro.au$^2$, uts.edu.au$^3$]
}

\begin{document}
\maketitle

\setlength{\abovedisplayskip}{3pt}
\setlength{\belowdisplayskip}{3pt}

\input{sections/abstract.tex}
\input{sections/introduction.tex}
\input{sections/preliminaries.tex}
\input{sections/method.tex}

\input{sections/connection_to_EM.tex}
\input{sections/experiments.tex}
\input{sections/conclusions.tex}

\bibliographystyle{named}
\let\oldbibliography\thebibliography
\renewcommand{\thebibliography}[1]{\oldbibliography{#1}
\setlength{\itemsep}{0pt}}
\bibliography{bibtex}

\input{sections/appendix.tex}

\end{document}

%% file: sections/abstract.tex
%
\begin{abstract}
  In this paper, we develop an efficient non-parametric Bayesian estimation of the kernel function of Hawkes processes. The non-parametric Bayesian approach is important because it provides flexible Hawkes kernels and quantifies their uncertainty.
  Our method is based on the cluster representation of Hawkes processes.
  Utilizing the finite support assumption of the Hawkes process, we efficiently sample random branching structures and thus, we split the Hawkes process into clusters of Poisson processes.
  We derive two algorithms --- a block Gibbs sampler and a maximum a posteriori estimator based on expectation maximization --- and we show that our methods have a linear time complexity, both theoretically and empirically.
  On synthetic data, we show our methods to be able to infer flexible Hawkes triggering kernels. On two large-scale Twitter diffusion datasets, we show that our methods outperform the current state-of-the-art in goodness-of-fit and that the time complexity is linear in the size of the dataset.
  We also observe that on diffusions related to online videos, the learned kernels reflect the perceived longevity for different content types such as music or pets videos.
\end{abstract}

%% file: sections/introduction.tex
%
\section{Introduction}
\label{sec:introduction}

\begin{table*}[t!]
\centering
\begin{tabular}{lcccc}
\toprule
Methods & Time Complexity & Bayesian & Continuous & Non-parametric \\
\midrule
\citet{zhou2013learning} & $O(n^3)$ & $\bm{\times}$ &    \checkmark        &  $\bm{\times}$ \\
\citet{xu2016learning} & $O(n^3)$ & $\bm{\times}$ &    \checkmark        &  $\bm{\times}$ \\
\citet{lewis2011nonparametric} & $O(n^3)$& $\bm{\times}$ &  $\bm{\times}$ & \checkmark \\
\citet{zhou2013learning_icml} & $O(n^3)$ & $\bm{\times}$ &  $\bm{\times}$ & \checkmark\\
\citet{rasmussen2013bayesian} & $O(n)$ & \checkmark &    \checkmark        &  $\bm{\times}$ \\
\citet{linderman2015scalable} &  $O(n)$      & \checkmark  &  \text{interval-censored} &    $\bm{\times}$            \\
\citet{rousseau2018nonparametric} &   unspecified  &    \checkmark      &   \checkmark      &    \checkmark            \\
Ours &  $O(n)$      & \checkmark  &  \checkmark   &    \checkmark\\
\bottomrule
\end{tabular}
\caption{Related Works on Non-parametric Hawkes Processes.} \label{tab:related_work}
\end{table*}

\begin{figure}[t!]
	\centering
	\subfloat[Poisson Cluster Process ]{{\includegraphics[scale=0.28, trim={5cm 2cm 11cm 2cm},clip]{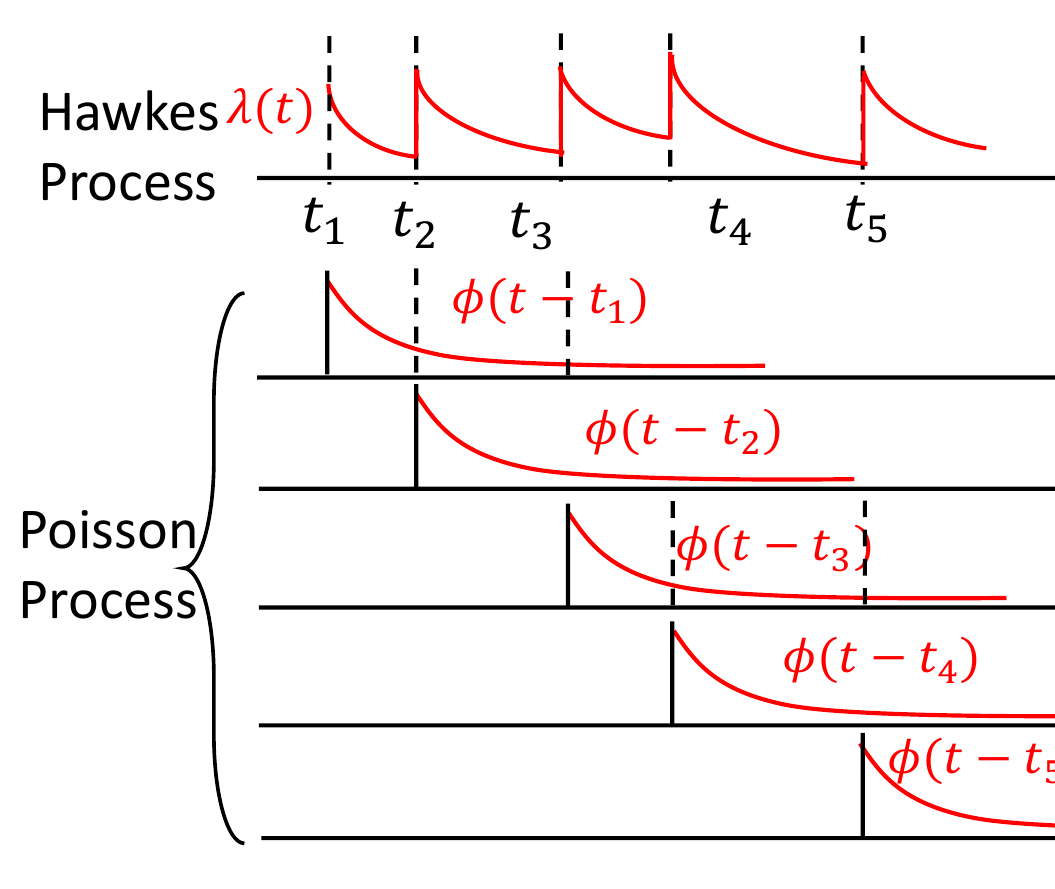}\label{subfig:poisson-cluster}}}
	\subfloat[Branching Structure]{\raisebox{4ex}{\includegraphics[scale=0.35, trim={11cm 6cm 13cm
				4cm},clip]{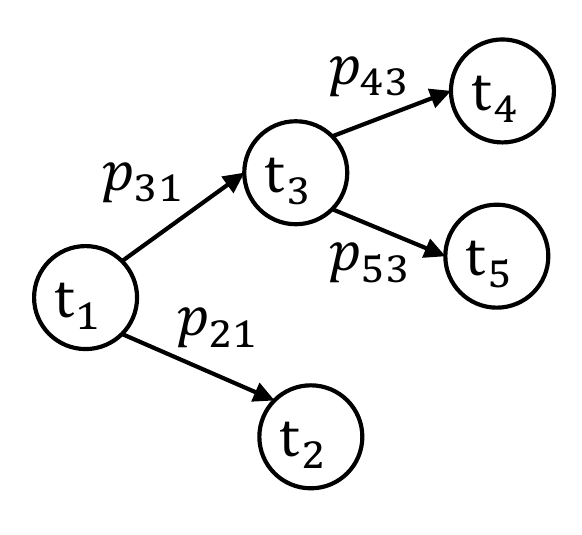}\label{subfig:branching-structure}}}
	\caption{
		The cluster Representation of a Hawkes Process.
		(a) A Hawkes process with decaying triggering kernel $\phi(\cdot)$ has intensity $\lambda(t)$ which increases after each new point is generated. It can be represented as a cluster of Poisson processes PP($\phi(t-t_i)$) associated with each $t_i$.
		(b) The branching structure corresponding to the triggering relationships shown in (a),
		where an edge $t_i \rightarrow t_j$ means that $t_i$ triggered $t_j$ with the probability $p_{ji}$ (calculated as \cref{eq:probability_branching_structure_phi}).
	}
	\label{fig:hp_2_pps}
\end{figure}

The Hawkes process \citep{10.2307/2334319} is a useful model of self-exciting point data in which the occurrence of a point increases the likelihood of arrival of new points.
More specifically, every point causes the conditional intensity function $\lambda$ --- which modulates the arrival rate of new points --- to increase.
An alternative representation of the Hawkes process is a cluster of Poisson processes \citep{10.2307/3212693}, which categorizes points into \textit{immigrants} and \textit{offspring}.
Immigrant points are generated independently at a background rate $\mu$; offspring points are triggered by existing points at a rate of $\phi$.
Points can therefore be structured into clusters,
where each cluster contains a point and the offspring it directly generated.
This leads to a tree structure, also known as
the branching structure (an example is shown in \cref{fig:hp_2_pps}).

\subsection{Background and Motivations.}
$\phi$ is important as it is shared and decides the class of the whole process and recently the Hawkes process with various $\phi$ has been studied.
\citet{Mishra2016FeaturePrediction} employ the branching factor of the Hawkes process with the power-law kernel to predict popularity of tweets;
\citet{kurashima2018modeling} predict human actions using a Hawkes process equipped with exponential, Weibull and Gaussian mixture kernels; online popularity unpredictability is explained using the Hawkes process with a variant of the exponential kernel by \citet{rizoiu2018sir}.
However, most work employes Hawkes process with parametric kernels,
which encodes strong assumptions, and limits the expressivity of the model.
Can we design a practical approach to learn flexible representations of the optimal Hawkes kernel function $\phi$ from data?

A typical solution is the non-parametric estimation of the kernel function \citep{lewis2011nonparametric,zhou2013learning_icml,bacry2014second}.
These are all frequentist methods which do not quantify the uncertainty of the learned kernels.
There exists work \citep{rasmussen2013bayesian,linderman2015scalable} on the Bayesian inference for the Hawkes process.
To scale past toy-dataset sizes
these methods require either parametric triggering kernels or discretization of the input domain, which in turn leads to poor scaling with the dimension of the domain and sensitivity to the choice of discretization.
The work closest to our own is that of \citet{rousseau2018nonparametric}, however their main contributions are theoretical; on the practical side they resort to an unscalable Markov chain Monte Carlo (MCMC) estimator.
We comparatively summarize related works in \cref{tab:related_work}.
To the best of our knowledge, our work is the first work proposing a Bayesian non-parametric Hawkes process estimation procedure,
with a linear time complexity allowing it to be applied to real-world datasets,
and without requiring discretization of domains.

\subsection{Contributions.}
In this paper, we propose a general framework for the efficient non-parametric Bayesian inference of Hawkes processes. 

(1) We exploit block Gibbs sampling \citep{ishwaran2001gibbs} to iteratively sample the latent branching structure, the background intensity $\mu$ and the triggering kernel $\phi$.
In each iteration, the point data are decomposed as a cluster of Poisson processes based on the sampled branching structure.
This is exemplified in \cref{fig:hp_2_pps}, in which a Hawkes process (shown on the top temporal axis of \cref{subfig:poisson-cluster}) is decomposed into several Poisson processes (the following temporal axes);
the corresponding branching structure is shown in \cref{subfig:branching-structure}.
The posterior $\mu$ and $\phi$ are estimated using the resulting cluster processes.
Our framework is close to the stochastic Expectation-Maximization (EM) algorithm \citep{celeux_diebolt85} where posterior $\mu$ and $\phi$ are estimated \citep{lloyd2015variational,pmlr-v70-walder17a} in the M-step and random samples of $\mu$ and $\phi$ are drawn.
We adapt the approach of the recent non-parametric Bayesian estimation for Poisson process intensities, termed Laplace Bayesian Poisson process (LBPP) \citep{pmlr-v70-walder17a}, to estimate the posterior $\phi$ given the sampled branching structure.

(2) We utilize the finite support assumption of the Hawkes Process to speed up sampling and computing the probability of the branching structure.
We theoretically show our method to be of linear time complexity.
Furthermore, we explore the connection with the EM algorithm \citep{Dempster77maximumlikelihood} and develop a second variant of our method, as an approximate EM algorithm.

(3) We empirically show our method enjoys linear time complexity and can infer known analytical kernels, i.e., exponential and sinusoidal kernels.
On two large-scale social media datasets, our method outperforms the current state-of-the-art non-parametric approaches and the learned kernels reflect the preceived longevity for different content types.

%% file: sections/preliminaries.tex
%
\section{Preliminaries}
In this section, we introduce the prerequisites of our work: the Hawkes process and LBPP.


\subsection{The Hawkes Process}

Introduced in \cref{sec:introduction}, the Hawkes process \citep{10.2307/2334319} can be specified using the conditional intensity function $\lambda$ which modulates the arrival rate of points. Mathematically, conditioned on a set of points
$\{t_i\}_{i=1}^{N}$, the intensity $\lambda$ is expressed as:
\begin{equation}
    \lambda(t) = \mu+\sum_{t_i<t} \phi(t-t_i), \label{eq:hawkes_lambda}
\end{equation}
where $\mu>0$, considered as a constant, and $\phi(\cdot):\mathbb{R}\rightarrow [0, \infty)$ are the background immigrant intensity and the triggering kernel.
The log-likelihood of $\{t_i\}_{i=1}^{N}$ given
$\mu$ and
$\phi(\cdot)$ is \citep{rubin1972regular}:
\begin{equation} \label{eq:log-likelihood}
\log p(\{t_i\}_{i=1}^{N} | \mu, \phi(\cdot))= \sum_{i=1}^{N} \log \lambda (t_i) -\int_{\Omega}\lambda(t) \intd t,
\end{equation}
where $\Omega$ is the sampling domain of $\{t_i\}_{i=1}^{N}$.

\subsection{Laplace Bayesian Poisson Process (LBPP)} \label{sec:lbpp}
LBPP \citep{pmlr-v70-walder17a} has been proposed for the non-parametric Bayesian estimation of the intensity of a Poisson process.
To satisfy non-negativity of the intensity function, LBPP models the intensity function $\lambda$ as a permanental process \citep{10.2307/3481499}, i.e.,  $\lambda =g \circ f $ where the link function $g(z)=z^2/2$ and $f(\cdot)$ obeys a Gaussian process (GP) prior.
Alternative link functions include $\exp(\cdot)$ \citep{moller1998log,diggle2013spatial} and $g(z) = \lambda^*(1+\exp(-z))^{-1}$ \citep{adams2009tractable} where $\lambda^*$ is constant.

The choice $g(z)=z^2/2$ has the analytical advantages; for some covariances the log-likelihood can be computed in closed form \citep{lloyd2015variational,flaxman2017poisson}.
LBPP exploits the Mercer expansion \citep{10.2307/91043} of the GP covariance function $k( x,  y) \equiv \text{Cov}(f( x), f(y))$,
\begin{equation}
\label{eq:mercer_expansion:general}
    k(x, y) = \sum_{i=1}^{K} \lambda_ie_i(x)e_i(y),
\end{equation}
where for non-degenerate kernels, $K=\infty$. The eigenfunctions $\{e_i(\cdot)\}_i$ are chosen to be orthonormal in $L^2(\Omega, m)$ for some sample space $\Omega$ with measure $m$. $f(\cdot)$ can be represented as a linear combination of $e_i(\cdot)$ \citep[section 2.2]{Rasmussen:2005:GPM:1162254},
$f(\cdot) = \bm{\omega}^T\bm{e(\cdot)}$,
and $\bm{\omega}$ has a Gaussian prior, i.e., $\bm{\omega} \sim \mathcal{N}(0, \Lambda)$ where $\Lambda=\diag(\lambda_1,\lambda_2,\cdots,\lambda_K)$ is a diagonal covariance matrix and $\bm{e}(\cdot) = [e_1(\cdot),\cdots,e_K(\cdot)]^T$ is a vector of basis functions. Computing the posterior distribution of the intensity function $\lambda(\cdot)$ is equivalent to estimating the posterior distribution of $\bm{\omega}$ which, in LBPP, is approximated by a normal distribution (a.k.a Laplace approximation \citep[section 3.4]{Rasmussen:2005:GPM:1162254}), i.e.,
\begin{align} \label{eq:laplace_approximate}
    \log p(\bm{\omega}|X, \Omega, k) \approx \log \mathcal{N}(\bm{\omega}|\hat{\bm{\omega}},Q),
\end{align}
where X$\equiv\{t_i\}_{i=1}^N$ is a set of point data, $\Omega$ the sample space and $k$ the Gaussian process kernel function. $\hat{\omega}$ is selected as the mode of the true posterior and Q the negative inverse Hessian of the true posterior at $\hat{w}$:
\begin{align}
    \hat{\bomega} =& \argmax_{\bm{\omega}}\log p(\bm{\omega}|X, \Omega, k), \label{eq:omega_est} \\
    Q^{-1} =& \left.-\partial_{\bomega \bomega^T} \log p(\bm{\omega}|X,\Omega,k)    \right\rvert_{\bomega=\hat{\bomega}}. \label{eq:Q_est}
\end{align}
The approximate posterior distribution of $f(t)$ is a normal distribution \citep[section 2.2]{Rasmussen:2005:GPM:1162254}:
\begin{equation}
\label{eq:posterior_f}
    f(t) \sim \mathcal{N}(\bm{\hat{\omega}}^T\bm{e}(t),\bm{e}(t)^T Q\bm{e}(t))\equiv \mathcal{N}(\nu, \sigma^2).
\end{equation}
Furthermore, the posterior of $\lambda(t)$ is a Gamma distribution:
\begin{equation}
\label{eq:posterior_phi}
	\GammaText(x|\alpha,\beta) \equiv\beta^{\alpha} x^{\alpha-1} e^{-\beta x}/\Gamma(\alpha),
\end{equation}
where $\alpha=(\nu^2+\sigma^2)^2/(4\nu^2\sigma^2+2\sigma^4)$ and $\beta = (\nu^2+\sigma^2)/(2\nu^2\sigma^2+\sigma^4)$.


%% file: sections/method.tex
%
\section{Inference via Sampling}
\label{sec:fast_bayesian_inference}

xWe now detail our efficient non-parametric Bayesian estimation algorithm, which samples the posterior of $\mu$ (constant background intensity) and $\phi(\cdot)$ (the trigerring kernel).
Our method starts with random $\mu_0, \phi_0(\cdot)$ and iterates by cycling through the following four steps ($k$ is the iteration index):
\renewcommand{\labelenumi}{\roman{enumi}}
\begin{enumerate}
  \item Calculate $p(\mathcal{B} |X,\phi_{k-1},\mu_{k-1})$, the distribution of the branching structure $\mathcal{B}$ given the data $X$, triggering kernel $\phi_{k-1}$, and background intensity $\mu_{k-1}$ (see \cref{sec:branching_structure}).
  \item Sample a $\mathcal{ B}_k$ as per $p(\mathcal{ B}|X,\phi_{k-1},\mu_{k-1})$ (see \cref{sec:branching_structure}).
  \item Estimate $p(\phi|\mathcal{B}_k,X)$ (\cref{sec:posterior_phi}) and $p(\mu|\mathcal{B}_k,X)$ (\cref{sec:posterior_mu}).
  \item Sample a $\phi_k(\cdot)$ and $\mu_k$ from $p(\phi(\cdot)|\mathcal{B}_k,X)$ and $p(\mu|\mathcal{B}_k,X)$, respectively.
\end{enumerate}
By standard Gibbs sampling arguments, the samples of $\phi(\cdot)$ and $\mu$ drawn in the step (iv) converge to the desired posterior, modulo the Laplace approximation in (iii). As the method is based on block Gibbs sampling \citep{ishwaran2001gibbs}, we term it \textit{Gibbs-Hawkes}.

\subsection{Distribution and Sampling of the Branching Structure} \label{sec:branching_structure}

The branching structure $\mathcal{B}$ has a data structure of tree (as \cref{fig:hp_2_pps}(b)) and consists of independent triggering events. 
Thus, the probability of the branching structure $\mathcal{ B}$ is the product of probabilities of triggering events, i.e., $p(\mathcal{ B})=\prod_{i=1}^{N}p_{ij_i}$ where $p_{ij_i}$ is the probability of $t_{j_i}$ triggering $t_i$.
Given $\mu$ and $\phi(\cdot)$, $p_{ij}$ is the ratio between $\phi(t_i-t_j)$ and $\lambda(t_i)$ (see e.g. \citep{lewis2011nonparametric}):
\begin{equation} \label{eq:probability_branching_structure_phi}
    p_{ij} \equiv \phi(t_i-t_j)/\lambda(t_i), ~~ j \geq 1.
\end{equation}
Similarly, the probability of point $t_i$ being from $\mu$ is:
\begin{equation}\label{eq:probability_branching_structure_mu}
p_{i0} \equiv \mu/\lambda(t_i) .
\end{equation}
Given these probabilities we may sample a branching structure by sampling a parent for each $t_i$ according to probabilities $\{p_{ij}\}_{j\geq0}$. The sampled branching structure separates a set of points into immigrants and offspring (introduced in \cref{sec:introduction}). Immigrants can be regarded as a sequence generated from PP($\mu$), where PP$(\mu)$ is a Poisson process which has an intensity $\mu$, and used to estimate the posterior $\mu$.

The key property which we exploit in the subsequent \cref{sec:posterior_mu} and \cref{sec:posterior_phi} is the following. Denote by $\{t_{k}^{(i)}\}_{k=1}^{N_{t_i}}$, the $N_{t_i}$ offspring generated by point $t_i$. If such a sequence is \textit{aligned} to an origin at $t_i$, yielding $S_{t_i}\equiv\{t_{k}^{(i)}-t_i\}_{k=1}^{N_i}$, then the aligned sequence is drawn from PP($\phi(\cdot)$) over [0, T-$t_i$] where $[0, T]$ is the sample domain of the Hawkes process. The posterior distribution of $\phi(\cdot)$ is estimated on all such aligned sequences.

\subsection{Posterior Distribution of $\mu$}
\label{sec:posterior_mu}
Continuing from the observations in \cref{sec:branching_structure}, note that if we are given a set of points $\{t_i\}_{i=1}^{M}$ generated by PP($\mu$) over $\Omega = [0, T]$, the likelihood for $\{t_i\}_{i=1}^{M}$ is the Poisson likelihood, $p(\{t_i\}_{i=1}^{M} |\mu, \Omega) =  e^{-\mu T}(\mu T)^M/M!$. For simplicity, we place a conjugate (Gamma) prior on $\mu T$, $\mu T \sim \GammaText(\alpha, \beta)$; the Gamma-Poisson conjugate family conveniently gives the posterior distribution of $\mu T$, \textit{i.e.}, $p(\mu T | \{t_i\}_{i=1}^{M}, \alpha, \beta)= \GammaText(\alpha+M, \beta+1)$.
We choose the scale $\alpha$ and the rate $\beta$ in the Gamma prior by making the mean of the Gamma posterior equal to $M$ and the variance $M/2$,
which is easily shown to correspond to $\alpha = M$ and $\beta= 1$ . Finally, due to conjugacy we obtain the posterior 
\begin{equation}
    p(\mu | \{t_i\}_{i=1}^{M}, \alpha, \beta)  = \GammaText(2M, 2T).
\end{equation}

\subsection{Posterior Distribution of $\phi$}
\label{sec:posterior_phi}

We handle the posterior distribution of the triggering kernel $\phi(\cdot)$ given the branching structure in an analogous manner to the LBPP method of \citet{pmlr-v70-walder17a}.
That is, we assume that $\phi(\cdot)=f^2(\cdot)/2$ where $f(\cdot)$ is Gaussian process distributed as described in \cref{sec:lbpp}. In line with \citep{pmlr-v70-walder17a}, we consider the sample domain $[0,\pi]$ and the so-called \textit{cosine kernel},
\begin{align}
    k(x,y)&=\sum_{\gamma \geq \bm{0}} \lambda_{\gamma}e_{\gamma}(x)e_{\gamma}(y), \label{eq:cos_kernel}
    \\
    \lambda_{\gamma}&\equiv 1/(a(
    \gamma^{2})^{m}+b), \label{eq:cos_lambda} 
    \\
    e_{\gamma}(x)&\equiv(2/\pi)^{1/2}\sqrt{1/2}^{[\gamma=0]}\cos{(\gamma x)}. \label{eq:cos_basis}
\end{align}
Here, $\gamma$ is a multi-index with non-negative (integral) values, $[\cdot]$ is the indicator function, $a$ and $b$ are parameters controlling the prior smoothness, and we let $m=2$. This basis is orthonormal w.r.t. the Lebesgue measure on $\Omega = [0,\pi]$. The expansion \cref{eq:cos_kernel} is an explicit kernel construction based on the Mercer expansion as per \cref{eq:mercer_expansion:general}, but other kernels may be used, for example by Nystr\"om approximation of the Mercer decomposition \citep{flaxman2017poisson}.

As mentioned at the end of \cref{sec:branching_structure}, by conditioning on the branching structure we may estimate $\phi(\cdot)$ by considering the \textit{aligned} sequences. In particular, letting $S_{t_i}$ denote the aligned sequence generated by $t_i$, the joint distribution of $\bomega$ and $\{S_{t_i}\}_{i=1}^{N}$ is calculated as \citep{pmlr-v70-walder17a}
{\medmuskip=1mu
\thinmuskip=1mu
\thickmuskip=1mu
\small
\begin{align}\label{eq:joint_distribution_for_data_and_omega}
    &\log p(\bomega, \{S_{t_i}\}_{i=1}^N | \Omega, k) \nonumber \\
    &= \sum_{i=1}^{N} \sum_{\Delta t \in S_{t_i}} \log \frac{1}{2} \left (\bomega^T\bm{e}(\Delta t) \right)^2-
    \dfrac{1}{2}\bomega^T (A+\Lambda^{-1}) \bomega+C,
    \\
    & A \equiv \sum_{i=1}^{N}\int_{0}^{T-t_i} \bm{e}(t)\bm{e}(t)^T
    \intd t,~~ C \equiv -\frac{1}{2}\log \left [(2\pi)^K |\Lambda| \right ]. \nonumber
\end{align}}where $K$ is the number of eigenfunctions.
Note that there is a subtle but important difference between the integral term above and that of \citet{pmlr-v70-walder17a}, namely the limit of integration; closed-form expressions for the present case are provided in Appendix \ref{sec:compute_integral}.
Putting the above equation into \cref{eq:omega_est} and \cref{eq:Q_est}, and we obtain the mean $\hat{\bomega}$ and the covariance $Q$ of the (Laplace) approximate log-posterior in $\bomega$:
{\medmuskip=0mu
	\thinmuskip=0mu
	\thickmuskip=0mu
	\small
\begin{align}
    \hat{\bm{\omega}} =& \argmax_{\bm{\omega}}  ~\log p(\bomega, \{S_{t_i}\}_{i=1}^N | \Omega, k),
    \label{eq:lbhp_omega} \\ 
    Q^{-1} =& -\sum_{i=1}^{N} \sum_{\Delta t \in S_{t_i}} 2\bm{e}(\Delta t)\bm{e}(\Delta t)^T/(\hat{\bomega}^T\bm{e}(\Delta t))^2 +A +\Lambda^{-1}. \label{eq:lbhp_Q}
\end{align}}Then, the posterior $\phi$ is achieved by \cref{eq:posterior_f,eq:posterior_phi}.

\subsection{Computational Complexity}
\label{sec:computational_complexity}
For LBPP, constructing \cref{eq:joint_distribution_for_data_and_omega} and \cref{eq:lbhp_Q}  takes $O(N_{o}K^2)$ where $K$ is the number of basis functions and $N_{o}$ is the number of offspring. Optimizing $\bomega$ (\cref{eq:lbhp_omega}) is a concave problem, which can be solved efficiently.
If L-BFGS is used, $O(CK)$ will be taken to calculate the gradient on each $\bomega$ where $C$ is the number of steps stored in memory. Computing $Q$ requires inverting a $K\times K$ matrix, which is $O(K^3)$.
As a result, the complexity of estimating $p(\phi | B)$ is $O((N_o+K)K^2)$. In terms of estimating $p(\mu | B)$ taking $O(1)$, the complexity of estimating $p(\mu|B)$, $p(\phi | B)$ is linear to the number of data. The time taken to sample $\mu$ and $\phi$ is minor ($O(1)$ and $O(K)$ respectively), so estimation time dominates.
While the naive complexity for $p_{ij}$ is $O(N^2)$, \citet{halpin2012algorithm} provides an optimized approach to reduce it to $O(N)$, which relies on the finite support assumption of Hawkes processes. \textbf{The finite support assumption} says that the value of the triggering kernel is negligible when the input is large \citep[p. 9]{halpin2012algorithm}.
As a result, the step of sampling branching structures can also be run in $O(N)$ and points with negligible impacts on another point are not sampled as its parents.
Interestingly, in comparison with LBPP, while our model is in some sense more complex, it enjoys a more favorable computational complexity.
In summary, we have the following complexities per iteration and in \cref{sec:experiments}, we validate the complexity on both synthetic and real data.
\begin{table}[b!]
	\centering
	\begin{tabular}{cc}
		\toprule
		Operation & Complexity \\
		\midrule
		$p(\mu|B)$ & $O(1)$ \\
		$p_{ij}$ & $O(N)$ \\
		$p(\phi|B)$ & $O((N_o+K)K^2)$ \\
		overall & $O((N+K)K^2)$ \\
		\bottomrule
	\end{tabular}
\caption{Time Complexity.}
\end{table}

%% file: sections/connection_to_EM.tex
%
\section{Maximum-A-Posterior Estimation}
\label{sec:relation_to_EM}

\begin{figure}[t] 
	\centering
	\includegraphics[width=1.03\columnwidth,trim={3.5cm 4cm 4cm 2.5cm},clip]{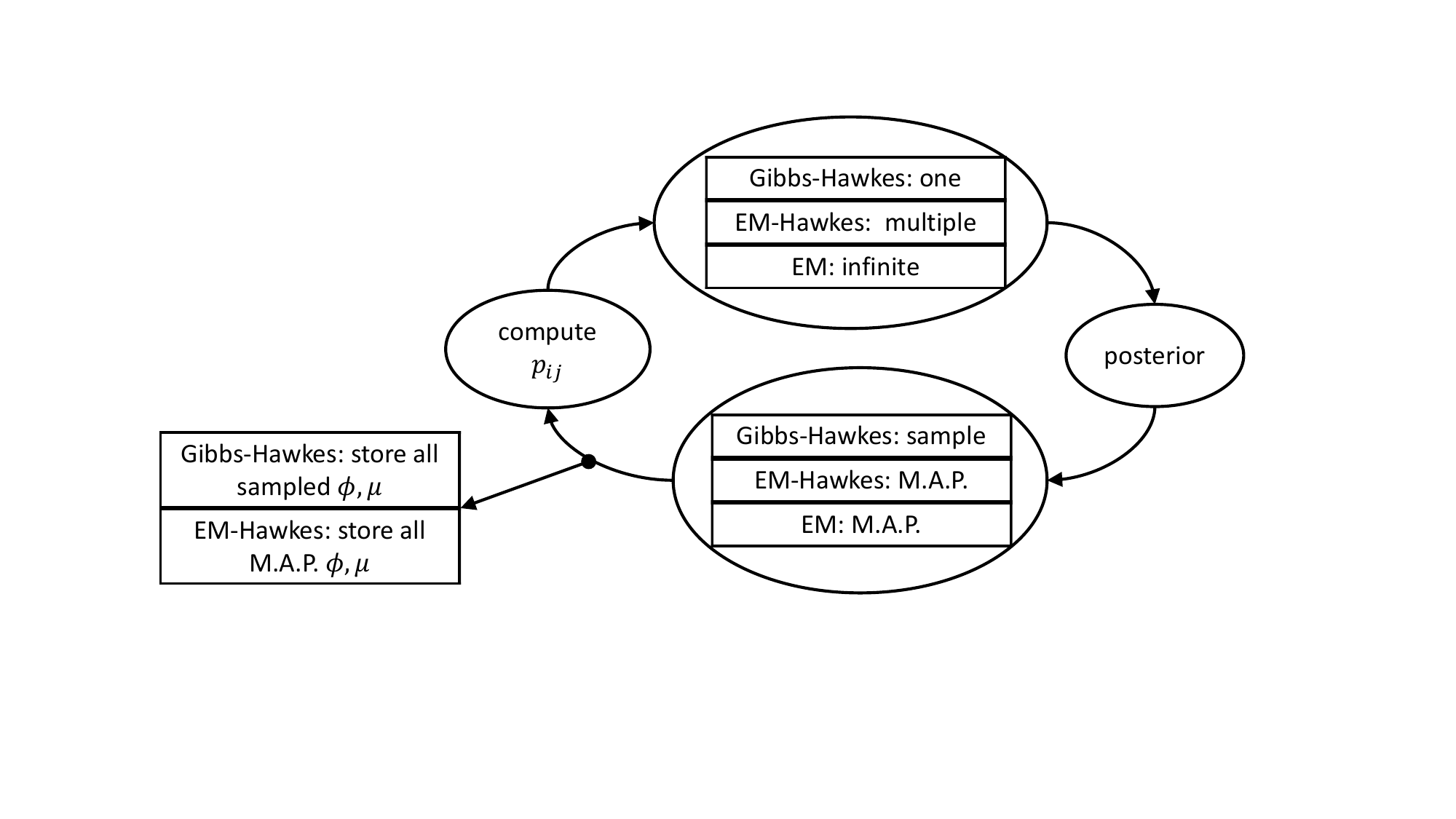}
	\caption{
		A visual summary of the Gibbs-Hawkes, EM-Hawkes and the EM algorithms. 
		The differences between them are 
		(1) the number of sampled branching structures and 
		(2) selected $\phi$ and $\mu$ for $p_{ij}$. 
		In contrast with with Gibbs-Hawkes, the EM-Hawkes method draws multiple branching structures at once and calculates $p_{ij}$ using M.A.P. $\phi$ and $\mu$. 
		The EM algorithm is equivalent to sampling infinite branching structures and exploiting M.A.P. or constrained M.L.E. $\phi$ and $\mu$ to calculate $p_{ij}$ (see \cref{sec:relation_to_EM}).
	}
	\label{fig:compare_three_methods}
\end{figure}
We explore a connection between the sampler of \autoref{sec:fast_bayesian_inference} and the EM algorithm, which allows us to introduce an analogous but intermediate scheme between them. In contrast to the random sampler of Section~\ref{sec:fast_bayesian_inference}, the proposed scheme  employs a deterministic \textit{maximum-a-posteriori} (M.A.P.) sampler.


\textbf{Relationship to EM.} 
In \cref{sec:introduction}, we mentioned the connection between our method and the stochastic EM algorithm \citep{celeux_diebolt85}. 
The difference is in the M-step; 
to perform EM \citep{Dempster77maximumlikelihood} we need only modify our sampler by: 
(a) sampling infinite branching structures at each iteration, and 
(b) re-calculating the probability of the branching structure with the M.A.P. $\mu$ and $\phi(\cdot)$, given the infinite set of branching structures.
More specifically, maximizing the expected log posterior distribution to estimate M.A.P. $\mu$ and $\phi(\cdot)$ given infinite branching structures is equivalent to maximizing the EM objective in the M-step (see 
Appendix \ref{sec:appx_map}
for the formal derivation).
Finally, note that the above step (b) 
is identical to the E-step of the EM algorithm.

\textbf{EM-Hawkes.}\label{sec:variant}
Following the discussion above, we propose \textit{EM-Hawkes}, an approximate EM algorithm
variant of Gibbs-Hawkes proposed in \cref{sec:fast_bayesian_inference}. 
Specifically, at each iteration
EM-Hawkes (a) samples a finite number of cluster assignments (to approximate the expected log posterior distribution), and 
(b) finds the M.A.P. triggering kernels and background intensities rather than sampling them as per block Gibbs sampling (the M-step of the EM algorithm). 
An overview of the Gibbs-Hawkes, EM-Hawkes and EM algorithm is illustrated in \cref{fig:compare_three_methods}. 

Note that under our LBPP-like posterior, finding the most likely triggering kernel $\phi(\cdot)$ is intractable 
(see Appendix \ref{sec:pos_distribution_multiple_phi_values})
As an approximation we take the element-wise mode of the \textit{marginals} of the $\{\phi(t_i)\}_{i}$ to approximate the mode of the joint distribution of the $\{\phi(t_i)\}_{i}$.

%% file: sections/experiments.tex
%

\newcommand\phicos{\phi_{\text{cos}}}
\newcommand\phiexp{\phi_{\text{exp}}}
\newcommand\mucos{\mu_{\text{cos}}}
\newcommand\muexp{\mu_{\text{exp}}}

\section{Experiments}

\label{sec:experiments}

\begin{figure}
	\centering
	\includegraphics[width=0.31\textwidth]{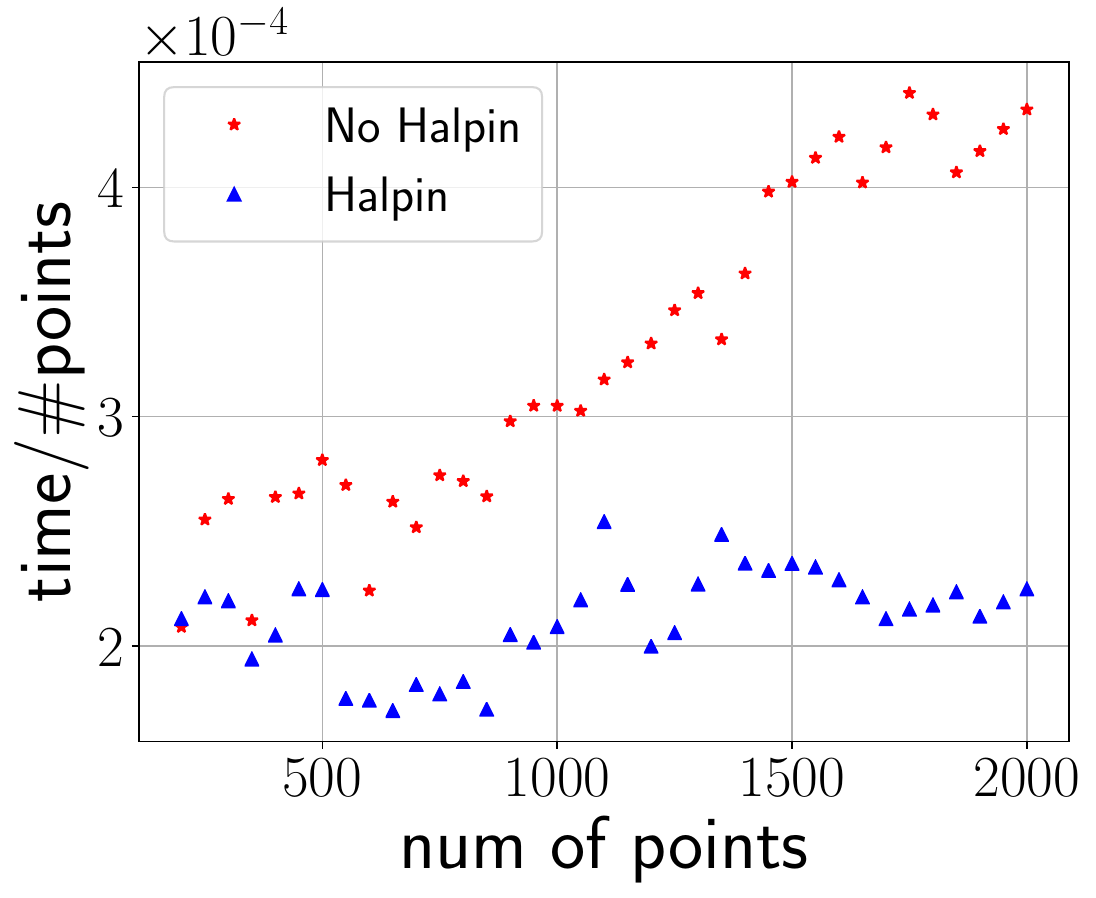}
	\caption{Computation time for calculating $p_{ij}$ and sampling branching structures, with and without Halpin's speed up.}
	\label{fig:halpin}
\end{figure}
\begin{figure}[t]
	\centering
	\includegraphics[width=0.31\textwidth]{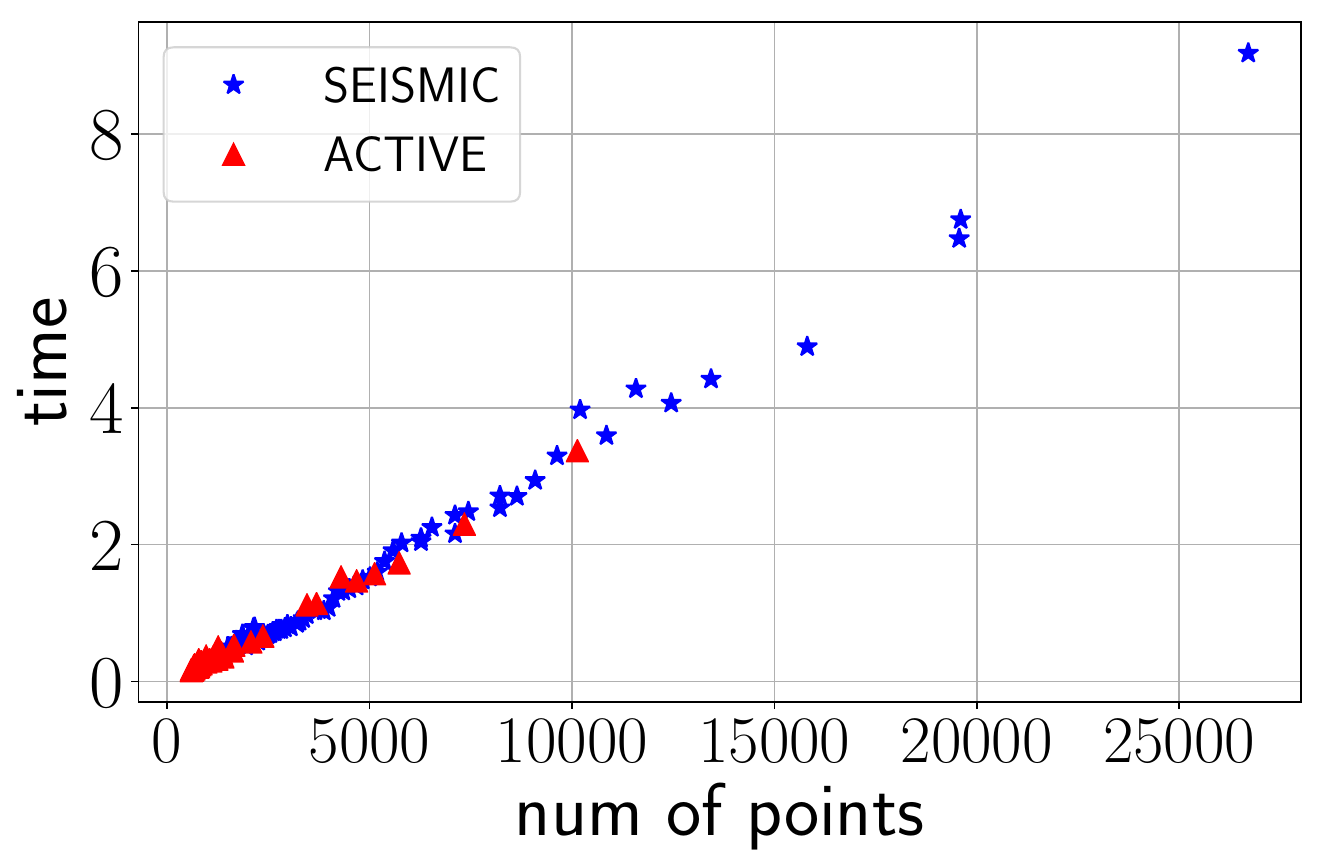}
	\caption{Running time per iteration on ACTIVE and SEISMIC.}
	\label{fig:run_time_real_date}
\end{figure}

We now evaluate our proposed approaches --- Gibbs-Hawkes and EM-Hawkes --- and compare them to three baseline models, on synthetic data and on two large Twitter online diffusion datasets.
The three baselines are: (1) A naive parametric Hawkes equipped with a constant background intensity and an exponential (Exp) triggering kernel $\phi = a_1a_2\exp(-a_2 t)$, $a_1, a_2 >0$, estimated by maximum likelihood.
(2) Ordinary differential equation (ODE)-based non-parametric non-bayesian Hawkes \citep{zhou2013learning_icml}.
(3) Wiener-Hopf (WH) equation based non-parametric non-bayesian Hawkes \citep{bacry2014second}. Codes of ODE based and WH based methods are publicly available \citep{2017arXiv170703003B}.

\subsection{Synthetic Data}
We employ two toy Hawkes processes to generate data, both having the same background intensity $\mu=10$, and cosine (\cref{eq:toy_phi_cos}) and exponential (\cref{eq:toy_phi_exp}) triggering kernels respectively. Note that compared to the cosine triggering kernel, the exponential one has a larger L2 norm for its derivative, and the difference is designed to test the performance of the approaches in different situations. We can check that both triggering kernels have negligible values when the input is large in the domain which we will choose as $[0, \pi]$, so the finite support assumption is satisfied.
{\medmuskip=1mu
	\thinmuskip=1mu
	\thickmuskip=1mu
\begin{align}
\phicos(t)=&
\cos(3\pi t)+1, \quad t\in (0,1]; \quad 0, \quad \text{otherwise}; \label{eq:toy_phi_cos}\\
\phiexp(t)=&5\exp(-5t), \quad t>0. \label{eq:toy_phi_exp}
\end{align}}
\vspace{-0.4cm}

\paragraph{Prediction.} 
For three baseline models and EM-Hawkes, the predictions $\mu_{\text{pred}}$ and $\phi_{\text{pred}}(\cdot)$ are taken to be the M.A.P. values, while for Gibbs-Hawkes we use the posterior mean.

\paragraph{Evaluation.} Each toy model generates 400 point sequences over $\Omega=[0,\pi]$, which are evenly split into 40 groups, 20 for training and 20 for test. 
Each of the three methods fit on each group, \textit{i.e.}, summing log-likelihoods for 10 sequences (for the parametric Hawkes) or estimating the log posterior probability of the Hawkes process given 10 sequences (for Gibbs-Hawkes and EM-Hawkes) or fitting the superposition of 10 sequences \citep{xu2017benefits}.
Since the true models are known, we evaluate fitting results using the relative L2 distance between predicted and true $\mu$ and $\phi(\cdot)$: $d_{L2}(g_{\text{pred}},g_{\text{true}}) = (\int_{\Omega} \big( g_{\text{pred}}(t)-g_{\text{true}}(t) \big )^2 dt)^{1/2}/(\int_{\Omega} (g_{\text{true}}(t) )^2 dt)^{1/2}$.

\begin{table}[tb]
	\begin{tabular}{llllll}
		\toprule
		Data          & Exp   & ODE   & WH & Gibbs & EM     \\
		\midrule
		$\phicos$  & 0.661 & 0.553 & 1.000 & 0.338 &0.318   \\
		$\mucos$   & 0.069 & 0.071  & 1.739 & 0.078 & 0.119 \\
		$\text{AVG}_\text{cos}$   & 0.365 & 0.312 & 1.370 & \textbf{0.208} & \underline{0.219} \\
		$\phiexp$  & 0.120 & 0.610 & 1.000 & 0.147& 0.140  \\
		$\muexp$   & 0.086 & 0.309 & 4.631 & 0.103 & 0.204 \\
		$\text{AVG}_\text{exp}$   & \textbf{0.103} & 0.460 & 2.816& \underline{0.125} & 0.172  \\
		\midrule
		ACTIVE   & 2.369 & 2.370 & 1.315 & \underline{2.580} &\textbf{2.592}   \\
		SEISMIC     & 3.335 & 3.357 & 2.131 & \underline{3.576} & \textbf{3.578} \\
		\bottomrule
	\end{tabular}
	\caption{Empirical performance comparison between algorithms (columns) with different measures (rows). \textit{Top:} relative L2 distance to known $\phi$ and $\mu$, and AVG denoted the average of L2 errors of $\phi$ and $\mu$. \textit{bottom:} mean predictive log likelihood on real data. Bold numbers denote the best performance and the underlined numbers for the second best.} 
\label{tab:results}
\end{table}

\paragraph{Experimental Details.}
For Gibbs-Hawkes and EM-Hawkes, we must select parameters of the GP kernel (\cref{eq:cos_kernel,eq:cos_lambda,eq:cos_basis}). 
An arbitrary choice of them can lead to poor performance, and to this end, we apply the standard cross validation based on the log-likelihood. We choose  the number of basis functions in $[8,16,32,64,128]$ and $a=b$ from $[0.2,0.02,\cdots,2\times10^{-8}]$.
We found that having many basis functions leads to a high fitting accuracy, but low speed. So, we use 32 basis functions which provides a suitable balance. 
In terms of kernel parameters $a,b$ of Equation~\eqref{eq:cos_lambda}, we observed that large values return smooth triggering kernels which have a large distance to the ground truth, while small values result in non-smooth predictions which however have small log-likelihoods. As a result, the values $a,b = 0.002$ were chosen. 5000 iterations are run to fit each group and first 1000 are ignored (i.e. \emph{burned-in}). 

\begin{figure*}[t]
	\centering
	\subfloat[Exponential synthetic data\label{fig:toy_exp}]{{\includegraphics[width=0.32\textwidth]{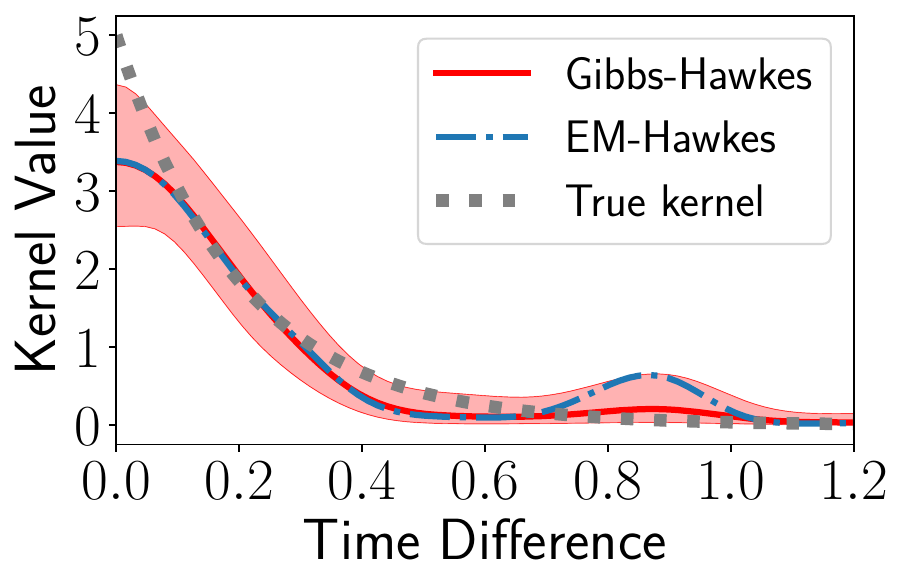} }}%
	\subfloat[ACTIVE vs. SEISMIC\label{fig:active_seismic}]{{\includegraphics[width=0.32\textwidth]{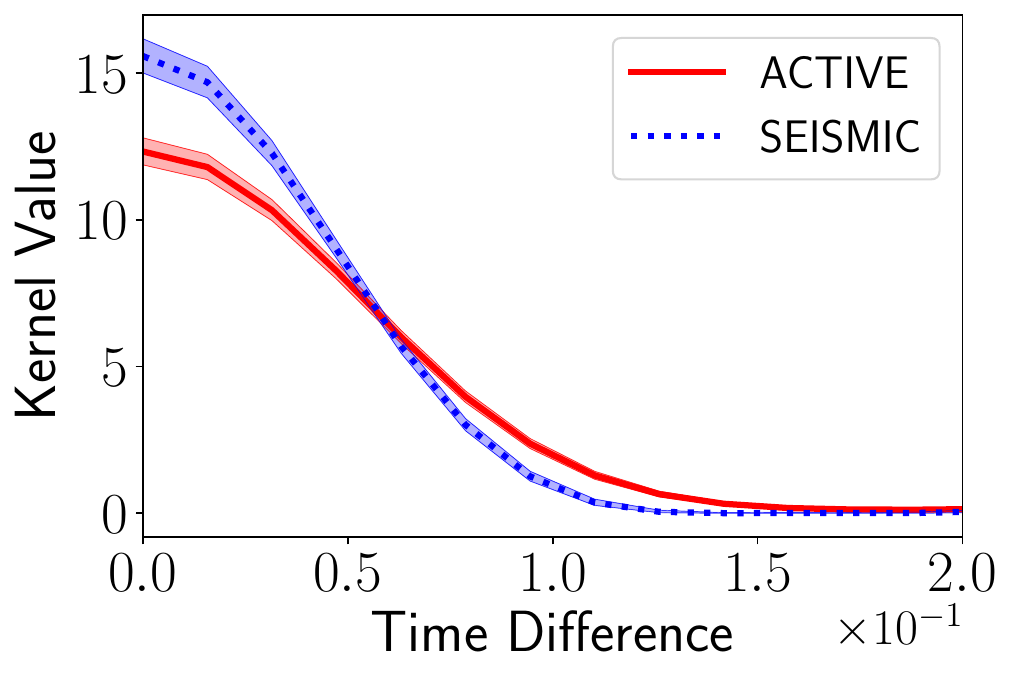}}}%
	\subfloat[Categories Music vs. Pets\label{fig:music_pets}]{{\includegraphics[width=0.32\textwidth]{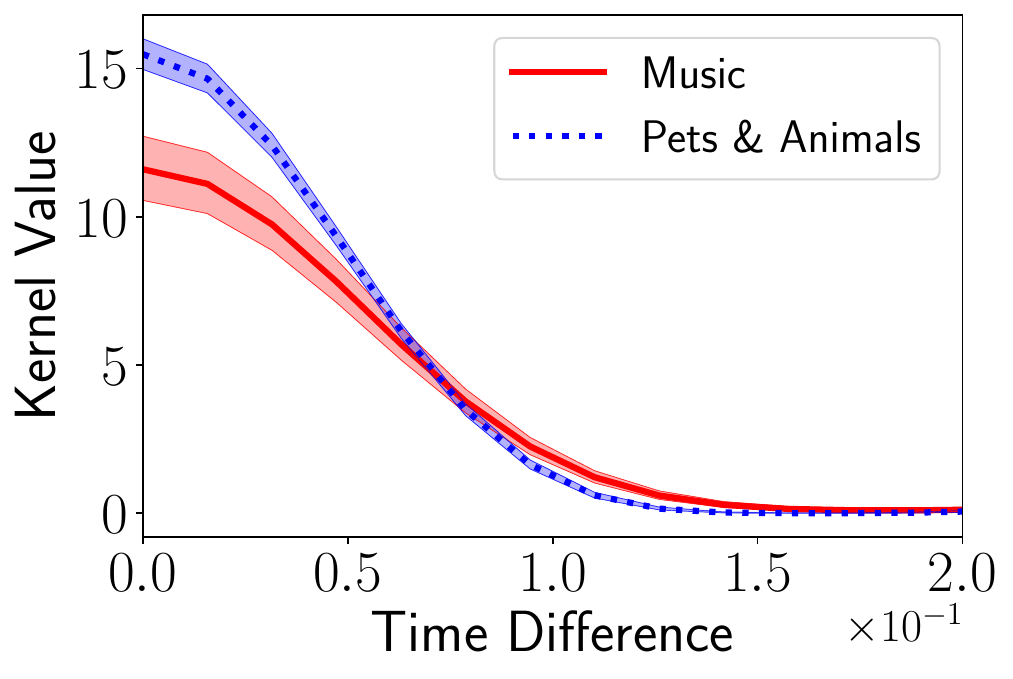} }}%
	\caption{
		Learned Hawkes triggering kernels using our non-parametric Bayesian approaches.
		Each red or blue area shows the estimated posterior distributions of $\phi$, while the solid lines indicate the $10$, $50$ and $90$ percentiles.
	    (a) A synthetic dataset simulated using $\phi_{\text{exp}}(t)$ (\cref{eq:toy_phi_exp}, shown in gray), is fit using Gibbs-Hawkes (in red) and EM-Hawkes (in blue);
	    (b) Twitter data in ACTIVE (in red) and SEISMIC (in blue);
	    (c) Twitter data associated with two categories in the ACTIVE set: Music (in red) and Pets \& Animals (in blue).
	}
\end{figure*}

\paragraph{Results.} 
The top of \cref{tab:results} shows the mean relative L2 distance between the learned and the true $\phi$ and $\mu$ on toy data.
First, Gibbs-Hawkes and EM-Hawkes are the closest the models to the ground truth cosine model according to the average error values ($\text{AVG}_{\text{cos}}$). For the exponential simulation model, both approaches gain the second and the third lowest errors respectively among all methods, and 
as expected, the parametric Hawkes -- which uses an exponential kernel -- fits the model best.
In contrast, the parametric model retrieves the cosine model worse because of its mismatch with the ground truth model.
The learned triggering kernels for $\phiexp$ and $\phicos$ by our approaches are shown in \cref{fig:toy_exp} 
and \cref{fig:toy_cos} in the appendix.
The ODE-based method performs unsatisfactorily on both simulation settings and it is observed that it performs better on $(\mucos,\phicos)$ than on $(\muexp,\phiexp)$. We explain the second observation as that the regularization of the ODE-based method encourages those triggering kernels that have small L2 norms for their derivatives and the derivative of $\phi_{\exp}(x)$ has a larger norm than that of $\phi_{\cos}(x)$.
Notably, tuning the hyper-parameters of the WH method is challenging, and \cref{tab:results} shows the best result obtained after a rather exhaustive experimentation. We speculate that the overall better performance of our approaches is due to the  regularization induced by the prior distributions and less difficult hyper-parameter selection. In addition, we also note that EM-Hawkes always performs better at discovering triggering kernels than Gibbs-Hawkes and this observation also holds on the real-life data. Thus, we conclude that generating multiple samples per iteration tends to improve modeling of the triggering kernels.
In summary, compared with state-of-the-art methods, our approaches achieve better performances for data generated by kernels from several parametric classes; as expected,
the parametric models are only effective for data generated from their own class. 

\paragraph{Effect of Halpin's Procedure.} \label{sec:scalability}
In \cref{sec:computational_complexity}, we show that using Halpin's procedure reduces the complexity of calculating $p_{ij}$ from quadratic to linear. 
We now empirically validate this speed up.
To distinguish between quadratic and linear complexity, we compute the ratio between running time and data size, as shown in \cref{fig:halpin}.
The ratio when using Halpin's procedure remains roughly constant as data size increases (the ratio increases linearly without the optimization), which implies that Halpin's procedure renders linear calculation of $p_{ij}$ and of branching structures. 
Later, we will show the linear complexity of our method on real data.

\subsection{Twitter Diffusion Data}

%


We evaluate the performance of our two proposed approaches (Gibbs-Hawkes and EM-Hawkes) on two Twitter datasets, containing retweet cascades.
A retweet cascade contains an original tweet, together with its direct and indirect retweets.
Current state of the art diffusion modeling approaches~\citep{zhao2015seismic,Mishra2016FeaturePrediction,rizoiu2018sir} are based on the self-exciting assumption: users get in contact with online content, and then diffuse it to their friends, therefore generating a cascading effect.
The two datasets we use have been employed in prior work and they are publicly available:
\begin{itemize}
    \item ACTIVE \citep{rizoiu2018sir} contains 41k retweet cascades, each containing at least 20 (re)tweets with links to Youtube videos.
    It was collected in 2014 and each Youtube video (and therefore each cascade) is associated with a Youtube category, e.g., \textit{Music} or \textit{News}.
    \item SEISMIC \citep{zhao2015seismic} contains 166k randomly sampled retweet cascades, collected in from Oct 7 to Nov 7, 2011. 
    Each cascade contains at least 50 tweets. 
\end{itemize}

\paragraph{Setup.}
The temporal extent of each cascade is scaled to $[0,\pi]$, and assigned to either training or test data with equal probability. 
We bundle together groups of 30 cascades of similar size, and we estimate one Hawkes process for each bundle.
Unlike for the synthetic dataset, for the retweet cascades dataset there is no \emph{true} Hawkes process to evaluate against.
Instead, we measure using log-likelihood how well the learned model generalizes to the test set.
We use the same hyper-parameters values as for the synthetic data. Finally, we follow the prior works on these cascade datasets \citep{zhao2015seismic,rizoiu2018sir} by setting the background intensity $\mu$ as $0$, because the cascade datasets contain only the information of triggering relationships.

\paragraph{Fitting Performance.}
For each dataset, we calculate the log-likelihood per event for each tweet cascade obtained by three baselines and our approaches (\cref{tab:results}).
Visibly, our proposed methods consistantly outperform baselines, with EM-Hawkes performing slightly better than Gibbs-Hawkes (by 0.5\% for ACTIVE and 0.06\% for SEISMIC).
This seems to indicate that online diffusion is influenced by factors not captured by the parametric kernel, therefore justifying the need to learn the Hawkes kernels non-parametrically. 
As mentioned in the synthetic data part, the WH-based method has a disadvantage of hard-to-tune hyper-parameters, which leads to the worst performance among all methods.

\paragraph{Scalability.}
To validate the linear complexity of our method, we record running time per iteration of Gibbs-Hawkes on ACTIVE and SEISMIC in \cref{fig:run_time_real_date}. The running time rises linearly with the number of points increasing, in line with the theoretical analysis. Linear complexity makes our method scalable and applicable on large datasets.

\paragraph{Interpretation.}
We show in \cref{fig:active_seismic} and \cref{fig:music_pets} the learned kernels for information diffusions.
We notice that the learned kernels appear to be decaying and long-tailed, in accordance with the prior literature.
\cref{fig:active_seismic} shows that the kernel learned on SEISMIC is decaying faster than the kernel learned on ACTIVE.
This indicates that non-specific (i.e. random) cascades have a faster decay than video-related cascades, presumably due to the fact that Youtube videos stay longer in the human attention.
This connection between the type of content and the speed of the decay seems further confirmed in \cref{fig:music_pets}, where we show the learned kernels for two categories in ACTIVE: \textit{Music} and \textit{Pets \& Animals}.
Cascades relating to \textit{Pets \& Animals} have a faster decaying kernel than \textit{Music}, most likely because Music is an ever-green content.

%% file: sections/conclusions.tex
\section{Conclusions}
In this paper, we provided the first non-parametric Bayesian inference procedure for the Hawkes process which requires no discretization of the input domain and enjoys a linear time complexity.
Our method iterates between two steps.
First, it samples the branching structure, effectively transforming the Hawkes process into a cluster of Poisson processes.
Next, it estimates the Hawkes triggering kernel using a non-parametric Bayesian estimation of the intensity of the cluster Poisson processes.
We provide both a full posterior sampler and an EM estimation algorithm based on our ideas.
We demonstrated our approach can infer flexible triggering kernels on simulated data.
On two large Twitter diffusion datasets, our method outperforms the state-of-the-art in held-out likelihood. Moreover, the learned non-parametric kernel reflects the intuitive longevity of different types of content. The linear complexity of our approach is corroborated on both the synthetic and real problems. The present framework is limited to the univariate unmarked Hawkes process and will be extended to marked multivariate Hawkes process.

\section*{Acknowledgements}
This research was supported in part by the Australian Government through the Australian Research Council's Discovery Projects funding scheme (project DP180101985).

%% file: sections/appendix.tex
%

\clearpage
\appendix
\onecolumn

Accompanying the submission \textit{Efficient Non-parametric Bayesian Hawkes Processes}.


\section{Computing the Integral Term of the Log-likelihood} \label{sec:compute_integral}
We consider $\Omega = [0, T]$, the background intensity $\mu$, the triggering kernel $\phi(\cdot)=1/2 f(\cdot)^2$, $f(\cdot)=\bomega^T \bm{e}(\cdot)$, and data $\{t_i\}_{i=1}^{N}$, and the integral term in the log-likelihood is calculated as below
\begin{align}
    &\text{Integral Term} \nonumber \\
    &= -\dfrac{1}{2} \sum_{i=1}^{N} \int_{0}^{T}f^2(t-t_i)dt \nonumber \\
    &= -\dfrac{1}{2} \sum_{i=1}^{N} \int_{0}^{T-t_i}[\sum_{k=1}^{K}\omega_k e_k(t)]^2dt \nonumber \\
    &= -\dfrac{1}{2} \sum_{i=1}^{N}\sum_{k=1}^{K}\sum_{k'=1}^{K} \omega_{k}\omega_{k'} \underbrace{\int_{0}^{T-t_i} e_k(t)e_{k'}(t)dt}_{U_{kk'}^{(i)}} \nonumber \\
    &= -\dfrac{1}{2} \sum_{i=1}^{N} \bomega^T U^{(i)} \bomega.
\end{align}

In our case, \cref{eq:cos_basis} has $d=1$, i.e., $\phi_{k}(x) = (2/\pi)^{1/2}\sqrt{1/2}^{[k-1=0]} \cos[(k-1) x]$, $k = 1,2,\cdots$. The matrix $U^{(i)}$ is calculated as below:
\begin{align}
U_{1,1}^{(i)} =& \int_{0}^{T-t_1} \dfrac{1}{\pi} dt = \dfrac{T-t_i}{\pi}, \nonumber \\
U_{k>1,1}^{(i)} =& U_{1,k>1}^{(i)}=\dfrac{\sqrt{2}}{\pi} \dfrac{\sin[(k-1)(t_m-t_i)]}{k-1}, \nonumber \\
U_{k,k(k>1)} =& \dfrac{1}{\pi} \Big \{T-t_i + \dfrac{\sin[2(k-1) (T-t_i)]}{2(k-1)}\Big \}, \nonumber \\
U_{k,k'(k\neq k')} =& \dfrac{1}{\pi} \Big\{\dfrac{\sin[(k-k')(T-t_i)]}{k-k'} + \dfrac{\sin[(k+k'-2)(T-t_i)]}{k+k'-2} \Big\}. \nonumber 
\end{align}

\section{M.A.P. $\mu$ and $\phi$ Given Infinite Branching Structures}
\label{sec:appx_map}
 M.A.P. $\mu$ and $\phi$ given Infinite branching structures is written as:
\begin{align}
    &\argmax_{\bomega, \mu}  \mathbb{E}_{B}[\log p(\bm{\omega}, \mu|B, \{t_i\}_{i=1}^{N}, \Omega, k)]\nonumber \\ 
    &= \argmax_{\bomega, \mu} \underbrace{\mathbb{E}_{B}[\log p(\{t_i\}_{i=1}^{N}|\bm{\omega},\mu, B,\Omega, k)]}_\text{Expected Log-likelihood} +\underbrace{\log p(\bomega)+\log p(\mu)}_\text{Constraints} \nonumber \\
    &= \argmax_{\bomega, \mu} \sum_{i=1}\big \{\sum_{t_j<t_i}p_{ij}\log\dfrac{1}{2}[\bomega^T \bm{e}(t_i - t_j)]^2-p_{i0}\log \mu -\dfrac{1}{2}\int_{0}^{T-t_i}[\bomega^T\bm{e}(t-t_i)]^2dt\big \}-(\beta + 1)\mu T \nonumber \\
    &\qquad \qquad -\dfrac{1}{2}
    \bomega^T \Lambda^{-1}\bomega - (\alpha-1)\log \mu,
\end{align}
where $B$ represents the branching structure, $p_{ij}$ the probabilities of triggering relationships shown as \cref{eq:probability_branching_structure_phi} and \cref{eq:probability_branching_structure_mu}, and $\alpha,\,\beta$ are parameters of the Gamma prior of $\mu T$.
The second line is obtained using Bayes' rule, which shows M.A.P. $\mu$ and $\phi$ given Infinite branching structures is equivalent to maximizing the constrained expected log-likelihood, i.e., the objective function for the M-step of the EM algorithm and the third line is an explicit expression of the second line.

\section{Mode-Finding the Triggering Kernel}
\label{sec:pos_distribution_multiple_phi_values}
Here we demonstrate in detail the computational challenges involved in finding the posterior mode with respect to the value of the triggering kernel at multiple point locations. Consider the triggering kernel $\phi(\cdot)=\dfrac{1}{2}f^2(\cdot)$ where $f(\cdot)$ is Gaussian process distributed. For a dataset $\{t_i\}_{i=1}^{N}$, $\bm{X}\equiv\{f(t_i)\}_{i=1}^{N} = \{X_i\}_{i=1}^{N}$ has a normal distribution, i.e., $\{f(t_i)\}_{i=1}^{N} \sim \mathcal{N}(\bm{m},\bm{\Sigma})$ where $\bm{m}$ and $\bm{\Sigma}$ are the mean and the covariance matrix. The distribution of $\bm{Y}\equiv\{\phi(t_i)\}_{i=1}^{N}=\{Y_i\}_{i=1}^{N}$ is derived as below where $F$ is the cumulative density function and $f$ the probabilistic density function.
\begin{align}
    & F_{\bm{Y}}(\bm{y}) \nonumber \\
    &= P(-\sqrt{2y_i} < X_i < \sqrt{2y_i},i=1,\cdots,N) \nonumber \\
    &= \int_{-\sqrt{2y_1}}^{\sqrt{2y_1}}\cdots\int_{-\sqrt{2y_N}}^{\sqrt{2y_N}} \dfrac{1}{\sqrt{(2\pi)^N \bm{\Sigma}^{-1}}}\exp[-\dfrac{(\bm{X}-\bm{m})^T \bm{\Sigma}^{-1}(\bm{X}-\bm{m}) }{2}] \intd X_1 \cdots \intd X_N, \nonumber\\
    &f_{\bm{Y}}(\bm{y}) \nonumber \\
    &=\dfrac{\partial^N}{\partial y_1\cdots \partial y_N} F_{\bm{Y}}(\bm{y}) \nonumber \\
    &= \dfrac{1}{\sqrt{(2\pi)^N \bm{\Sigma}^{-1}}}(\Pi_{i=1}^{N} \dfrac{1}{2\sqrt{2y_1}})\sum_{\bm{X} \in \bigtimes_{i=1}^{N} \{\sqrt{2y_1},-\sqrt{2y_1}\}}\exp[-\dfrac{(\bm{X}-\bm{m})^T \bm{\Sigma}^{-1}(\bm{X}-\bm{m}) }{2}],
\end{align}
where $\bigtimes$ is the Cartesian product.
There are $2^N$ summations of exponential functions, which is intractable.

\begin{figure*}[h]
\centering
\includegraphics[scale = 0.42]{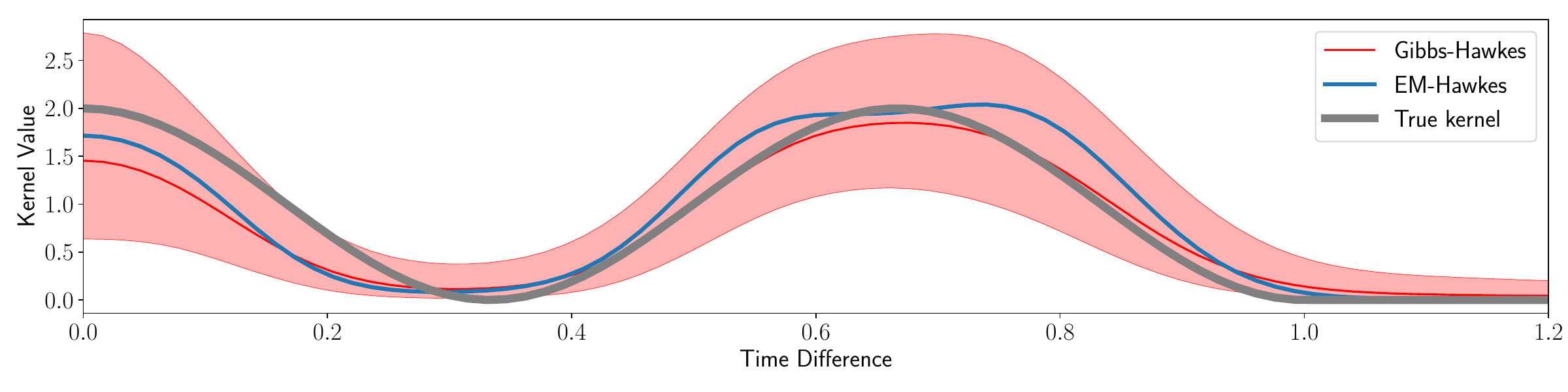}
\caption{Triggering kernels estimated by the Gibbs-Hawkes method (\cref{sec:fast_bayesian_inference}) and the EM-Hawkes method (\cref{sec:variant}). The true  kernel is plotted as the bold \textbf{\textcolor{gray}{gray}} curve. We plot the median (\textcolor{red}{red}) and [0.1, 0.9] interval (filled \textcolor{red}{red}) of the approximate predictive distribution, along with the triggering kernel inferred by the EM Hawkes method (\textcolor{blue}{blue}). The hyper-parameters $a$ and $b$ of the Gaussian process kernel (\cref{eq:cos_lambda}) are set to 0.002.}
\label{fig:toy_cos}
\end{figure*}


%% file: main.bbl
\begin{thebibliography}{}

\bibitem[\protect\citeauthoryear{Adams \bgroup \em et al.\egroup
  }{2009}]{adams2009tractable}
Ryan~Prescott Adams, Iain Murray, and David J~C MacKay.
\newblock Tractable nonparametric bayesian inference in poisson processes with
  gaussian process intensities.
\newblock In {\em ICML}, pages 9--16, 2009.

\bibitem[\protect\citeauthoryear{Bacry and Muzy}{2014}]{bacry2014second}
Emmanuel Bacry and Jean-Francois Muzy.
\newblock Second order statistics characterization of hawkes processes and
  non-parametric estimation.
\newblock {\em arXiv preprint}, 2014.

\bibitem[\protect\citeauthoryear{Bacry \bgroup \em et al.\egroup
  }{2017}]{2017arXiv170703003B}
Emmanuel Bacry, Martin Bompaire, Stéphane Gaïffas, and Soren Poulsen.
\newblock tick: a python library for statistical learning, with a particular
  emphasis on time-dependent modeling.
\newblock {\em ArXiv preprint}, 2017.

\bibitem[\protect\citeauthoryear{Celeux and Diebolt}{1985}]{celeux_diebolt85}
Gilles Celeux and Jean Diebolt.
\newblock The sem algorithm: A probabilistic teacher algorithm derived from the
  {EM} algorithm for the mixture problem.
\newblock {\em Comp. Stat. Quarterly}, 2:73--82, 1985.

\bibitem[\protect\citeauthoryear{Dempster \bgroup \em et al.\egroup
  }{1977}]{Dempster77maximumlikelihood}
Arthur~P Dempster, Nan~M Laird, and Donald~B Rubin.
\newblock Maximum likelihood from incomplete data via the em algorithm.
\newblock {\em Journal of the Royal Statistical Society, Series B},
  39(1):1--38, 1977.

\bibitem[\protect\citeauthoryear{Diggle \bgroup \em et al.\egroup
  }{2013}]{diggle2013spatial}
Peter~J Diggle, Paula Moraga, Barry Rowlingson, and Benjamin~M Taylor.
\newblock Spatial and spatio-temporal log-gaussian cox processes: extending the
  geostatistical paradigm.
\newblock {\em Statistical Science}, 28(4):542--563, 2013.

\bibitem[\protect\citeauthoryear{Donnet \bgroup \em et al.\egroup
  }{2018}]{rousseau2018nonparametric}
Sophie Donnet, Vincent Rivoirard, and Judith Rousseau.
\newblock Nonparametric bayesian estimation of multivariate hawkes processes.
\newblock {\em arXiv preprint}, 2018.

\bibitem[\protect\citeauthoryear{Flaxman \bgroup \em et al.\egroup
  }{2017}]{flaxman2017poisson}
Seth Flaxman, Yee~Whye Teh, and Dino Sejdinovic.
\newblock Poisson intensity estimation with reproducing kernels.
\newblock In {\em AISTATS}, pages 270--279, 2017.

\bibitem[\protect\citeauthoryear{Halpin}{2013}]{halpin2012algorithm}
Peter~F Halpin.
\newblock A scalable em algorithm for hawkes processes.
\newblock {\em New Developments in Quantitative Psychology}, pages 403--414,
  2013.

\bibitem[\protect\citeauthoryear{Hawkes and Oakes}{1974}]{10.2307/3212693}
Alan Hawkes and David Oakes.
\newblock A cluster process representation of a self-exciting process.
\newblock {\em Journal of Applied Probability}, 11(3):493--503, 1974.

\bibitem[\protect\citeauthoryear{Hawkes}{1971}]{10.2307/2334319}
Alan Hawkes.
\newblock Spectra of some self-exciting and mutually exciting point processes.
\newblock {\em Biometrika}, 58(1):83--90, 1971.

\bibitem[\protect\citeauthoryear{Ishwaran and James}{2001}]{ishwaran2001gibbs}
Hemant Ishwaran and Lancelot~F James.
\newblock Gibbs sampling methods for stick-breaking priors.
\newblock {\em Journal of the American Statistical Association},
  96(453):161--173, 2001.

\bibitem[\protect\citeauthoryear{Kurashima \bgroup \em et al.\egroup
  }{2018}]{kurashima2018modeling}
Takeshi Kurashima, Tim Althoff, and Jure Leskovec.
\newblock Modeling interdependent and periodic real-world action sequences.
\newblock In {\em WWW}, pages 803--812, 2018.

\bibitem[\protect\citeauthoryear{Lewis and
  Mohler}{2011}]{lewis2011nonparametric}
Erik Lewis and George Mohler.
\newblock A nonparametric em algorithm for multiscale hawkes processes.
\newblock {\em Journal of Nonparametric Statistics}, 1(1):1--20, 2011.

\bibitem[\protect\citeauthoryear{Linderman and
  Adams}{2015}]{linderman2015scalable}
Scott~W Linderman and Ryan~P Adams.
\newblock Scalable bayesian inference for excitatory point process networks.
\newblock {\em arXiv preprint}, 2015.

\bibitem[\protect\citeauthoryear{Lloyd \bgroup \em et al.\egroup
  }{2015}]{lloyd2015variational}
Chris Lloyd, Tom Gunter, Michael~A Osborne, and Stephen~J Roberts.
\newblock Variational inference for gaussian process modulated poisson
  processes.
\newblock In {\em ICML}, pages 1814--1822, 2015.

\bibitem[\protect\citeauthoryear{Mercer}{1909}]{10.2307/91043}
James Mercer.
\newblock Functions of positive and negative type, and their connection with
  the theory of integral equations.
\newblock {\em Philosophical Transactions of the Royal Society A},
  209:415--446, 1909.

\bibitem[\protect\citeauthoryear{Mishra \bgroup \em et al.\egroup
  }{2016}]{Mishra2016FeaturePrediction}
Swapnil Mishra, Marian-Andrei Rizoiu, and Lexing Xie.
\newblock Feature driven and point process approaches for popularity
  prediction.
\newblock {\em CIKM}, pages 1069--1078, 2016.

\bibitem[\protect\citeauthoryear{Møller \bgroup \em et al.\egroup
  }{1998}]{moller1998log}
Jesper Møller, Anne~Randi Syversveen, and Rasmus~Plenge Waagepetersen.
\newblock Log gaussian cox processes.
\newblock {\em Scandinavian journal of statistics}, 25(3):451--482, 1998.

\bibitem[\protect\citeauthoryear{Rasmussen and
  Williams}{2005}]{Rasmussen:2005:GPM:1162254}
Carl~Edward Rasmussen and Christopher K~I Williams.
\newblock {\em Gaussian Processes for Machine Learning}.
\newblock The MIT Press, 2005.

\bibitem[\protect\citeauthoryear{Rasmussen}{2013}]{rasmussen2013bayesian}
Jakob~Gulddahl Rasmussen.
\newblock Bayesian inference for hawkes processes.
\newblock {\em Methodology and Computing in Applied Probability},
  15(3):623--642, 2013.

\bibitem[\protect\citeauthoryear{Rizoiu \bgroup \em et al.\egroup
  }{2018}]{rizoiu2018sir}
Marian-Andrei Rizoiu, Swapnil Mishra, Quyu Kong, Mark Carman, and Lexing Xie.
\newblock Sir-hawkes: Linking epidemic models and hawkes processes to model
  diffusions in finite populations.
\newblock In {\em WWW}, pages 419--428, 2018.

\bibitem[\protect\citeauthoryear{Rubin}{1972}]{rubin1972regular}
Izhak Rubin.
\newblock Regular point processes and their detection.
\newblock {\em IEEE Transactions on Information Theory}, 18(5):547--557, 1972.

\bibitem[\protect\citeauthoryear{Shirai and Takahashi}{2003}]{10.2307/3481499}
Tomoyuki Shirai and Yoichiro Takahashi.
\newblock Random point fields associated with certain fredholm determinants ii:
  Fermion shifts and their ergodic and gibbs properties.
\newblock {\em The Annals of Probability}, 31(3):1533--1564, 2003.

\bibitem[\protect\citeauthoryear{Walder and Bishop}{2017}]{pmlr-v70-walder17a}
Christian~J Walder and Adrian~N Bishop.
\newblock Fast bayesian intensity estimation for the permanental process.
\newblock In {\em ICML}, pages 3579--3588, 2017.

\bibitem[\protect\citeauthoryear{Xu \bgroup \em et al.\egroup
  }{2016}]{xu2016learning}
Hongteng Xu, Mehrdad Farajtabar, and Hongyuan Zha.
\newblock Learning granger causality for hawkes processes.
\newblock In {\em ICML}, pages 1717--1726, 2016.

\bibitem[\protect\citeauthoryear{Xu \bgroup \em et al.\egroup
  }{2018}]{xu2017benefits}
Hongteng Xu, Dixin Luo, Xu~Chen, and Lawrence Carin.
\newblock Benefits from superposed hawkes processes.
\newblock {\em AISTATS}, pages 623--631, 2018.

\bibitem[\protect\citeauthoryear{Zhao \bgroup \em et al.\egroup
  }{2015}]{zhao2015seismic}
Qingyuan Zhao, Murat~A Erdogdu, Hera~Y He, Anand Rajaraman, and Jure Leskovec.
\newblock Seismic: A self-exciting point process model for predicting tweet
  popularity.
\newblock In {\em SIGKDD}, pages 1513--1522, 2015.

\bibitem[\protect\citeauthoryear{Zhou \bgroup \em et al.\egroup
  }{2013a}]{zhou2013learning}
Ke~Zhou, Hongyuan Zha, and Le~Song.
\newblock Learning social infectivity in sparse low-rank networks using
  multi-dimensional hawkes processes.
\newblock In {\em AISTATS}, pages 641--649, 2013.

\bibitem[\protect\citeauthoryear{Zhou \bgroup \em et al.\egroup
  }{2013b}]{zhou2013learning_icml}
Ke~Zhou, Hongyuan Zha, and Le~Song.
\newblock Learning triggering kernels for multi-dimensional hawkes processes.
\newblock In {\em ICML}, pages III--1301--III--1309, 2013.

\end{thebibliography}
